\documentclass{article}
\usepackage{ijcai18}

\usepackage[labelfont=bf,small]{caption}
\usepackage{times}
\usepackage{xcolor}
\usepackage{soul}
\usepackage{url}
\usepackage[utf8]{inputenc}
\usepackage{caption}
\usepackage{latexsym}
\usepackage{amsmath}
\usepackage{amsfonts}
\usepackage{graphicx}
\usepackage{booktabs}
\usepackage{multirow}
\usepackage{comment}
\usepackage[T1]{fontenc}
\usepackage[linesnumbered,ruled,vlined]{algorithm2e}
\DeclareUnicodeCharacter{FF0C}{ }

\title{Scheduled Policy Optimization for Natural Language Communication with Intelligent Agents}

\author{
 Wenhan Xiong$^1$, 
 Xiaoxiao Guo$^2$, 
 Mo Yu$^2$, 
 Shiyu Chang$^2$, 
 Bowen Zhou$^3$, 
 William Yang Wang$^1$, 
\\ 
 $^1$ University of California, Santa Barbara\\
 $^2$ IBM Research\\
 $^3$ JD AI Research\\
 \{xwhan,william\}@cs.ucsb.edu 
 }
\begin{document}

\maketitle

\begin{abstract}
We investigate the task of learning to follow natural language instructions by jointly reasoning with visual observations and language inputs. In contrast to existing methods which start with learning from demonstrations (LfD) and then use reinforcement learning (RL) to fine-tune the model parameters, we propose a novel policy optimization algorithm which dynamically schedules demonstration learning and RL. The proposed training paradigm provides efficient exploration and better generalization beyond existing methods. Comparing to existing ensemble models, the best single model based on our proposed method tremendously decreases the execution error by over $50\%$ on a block-world environment. To further illustrate the exploration strategy of our RL algorithm, We also include systematic studies on the evolution of policy entropy during training.
\end{abstract}

\section{Introduction}
\label{sec:intro}
Language is a natural form for humans to express their intention. In recent years, although researchers have successfully built intelligent systems which are able to accomplish complicated tasks~\cite{levine2016learning,silver2017mastering}, few of them are able to cooperate with humans via natural language. To build better AI systems that can safely and robustly work along with people, it is necessary to teach machines to understand free-form human language instructions and output low-level working actions. This is a challenging task, mainly due to the ambiguity of human language and the complexity of the working environment.

\begin{figure}[t]
\centering
\includegraphics[width=0.9\linewidth]{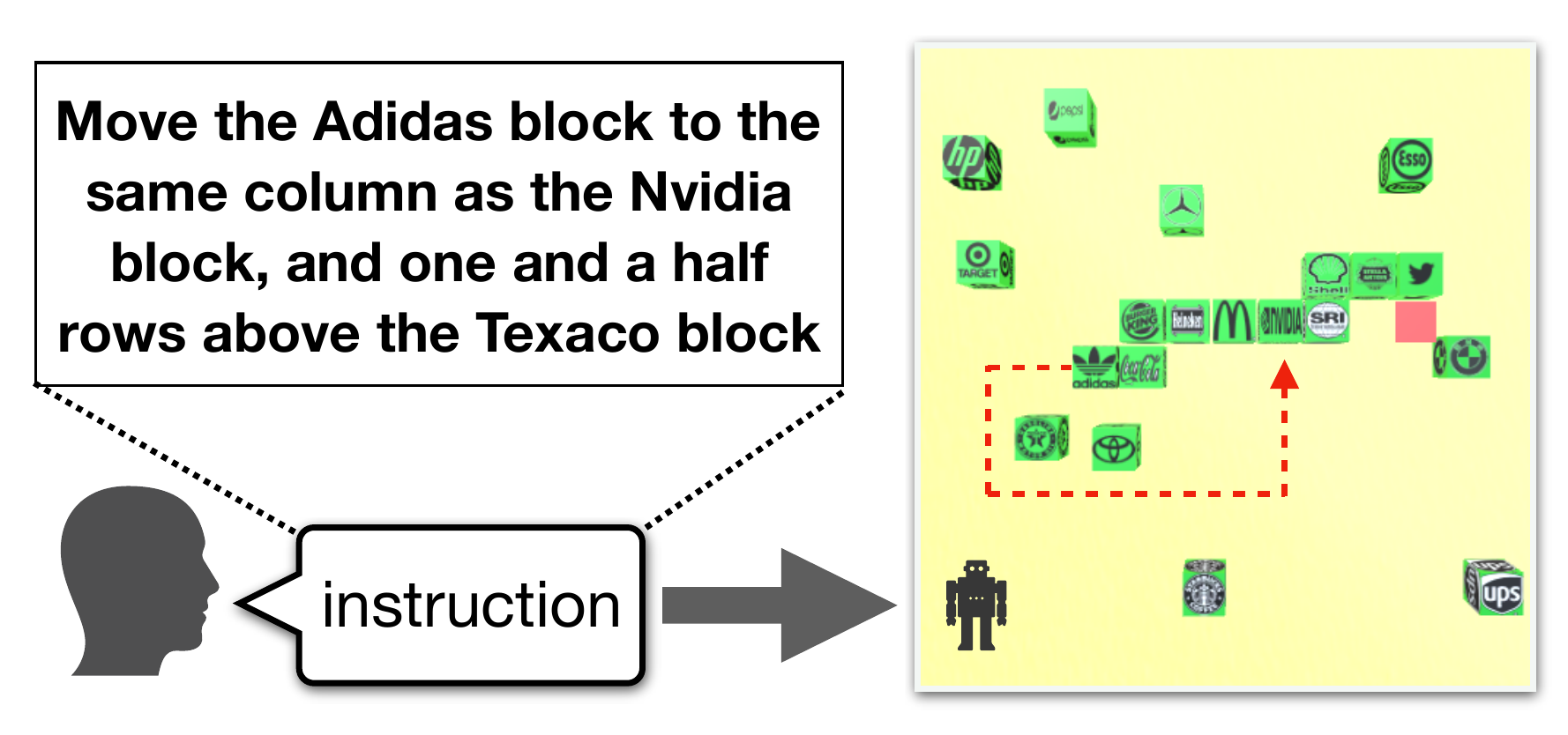}
\caption{Task illustration. The intelligent agent is expected to understand human language instructions and make sequential actions based on its observations about the working environment.}
\label{task}
\end{figure}

In this work, we aim at developing an intelligent agent which can take as inputs human language instructions as well as environment observations to finish the task specified by the human language in a simulated working environment~\cite{bisk2016natural,misra2017mapping}. The specific task is illustrated in Figure~\ref{task}. In order to accomplish the task, the agent should be able to recognize potential obstacles in the environment and move around. Besides, since the same task may be described by different humans, the agent must also be robust to language variations.

Early methods for similar tasks~\cite{chen2011learning,matuszek2010following,tellex2011understanding} rely on human defined spatial or language features to parse the language. Meticulous engineering in terms of environment domain and language lexicon is often required. In this work, we focus on developing a neural-network based model that can be trained end-to-end with minimum domain and linguistic knowledge.

More recently, the task of mapping natural language into low-level actions or programs has been tackled with neural network based methods~\cite{mei2016listen,liang2016neural}. In the simplest case, a cross-entropy loss can be used to train the model so that it can imitate the human-demonstrated actions. However, the pure supervised model fails to explore the state-action space outside the demonstration path, which undermines the model's generalization ability. 

To develop a model that is able to not only imitate but also generalize, Misra \emph{et al.}~\shortcite{misra2017mapping} apply various deep reinforcement learning (RL) techniques to this task. The RL agent is able to explore more state-action space via its stochastic policy (probability distribution over actions). Since RL from scratch can be highly data-inefficient due to sparse rewards and the large action space. Misra \emph{el al.}~\shortcite{misra2017mapping} warm-start the network parameters with several epochs of supervised learning which imitates human actions. The RL algorithm is then adopted to fine-tune the parameters. This training paradigm is successful at speeding up training. However, we show by experiments that the supervised pre-training often results in a high-entropy policy. When the agent samples actions from the high-entropy policy, the agent tends to make near-greedy decisions. This actually prevents the agent from exploring the consequences of choosing other actions. Their experiment results also indicate that there is still a large performance gap between humans and existing systems. 

In contrast to this training paradigm, we propose a novel scheduled policy optimization mechanism inspired by scheduled sampling~\cite{bengio2015scheduled}, which addresses the discrepancy between training and inference in sequence decoding. Our scheduling mechanism dynamically alternates between imitating the human actions (learning from demonstration) and reinforcement learning. Ideally, at the early stage of training, the scheduler should more frequently utilize demonstration learning to alleviate the sparse reward issue; as the agent acquires more experience, more RL updates should be scheduled to achieve better generalization. Empirically, we achieve the best performance on the block-world task, reducing the execution error by more than $50\%$, which is much closer to human performance. In summary, our main contributions are:
\begin{itemize}
	\item Based on the Block environment, we build a state-of-the-art system which is able to accomplish tasks described by free-form text.
	\item We propose a novel scheduled RL algorithm which achieves better data efficiency while maintaining sufficient exploration.
	\item We conduct systematic studies to compare the exploration strategies of different RL systems using the Block environment.
\end{itemize}

Our paper is organized as follows: we describe the proposed approach in Section~\ref{sec:method}. Experiment results and analysis are shown in Section~\ref{sec:exp}. We then discuss related work in Section~\ref{sec:related}. Finally, we conclude in Section~\ref{sec:con}.

\begin{figure*}[t]
\includegraphics[width=0.90\textwidth]{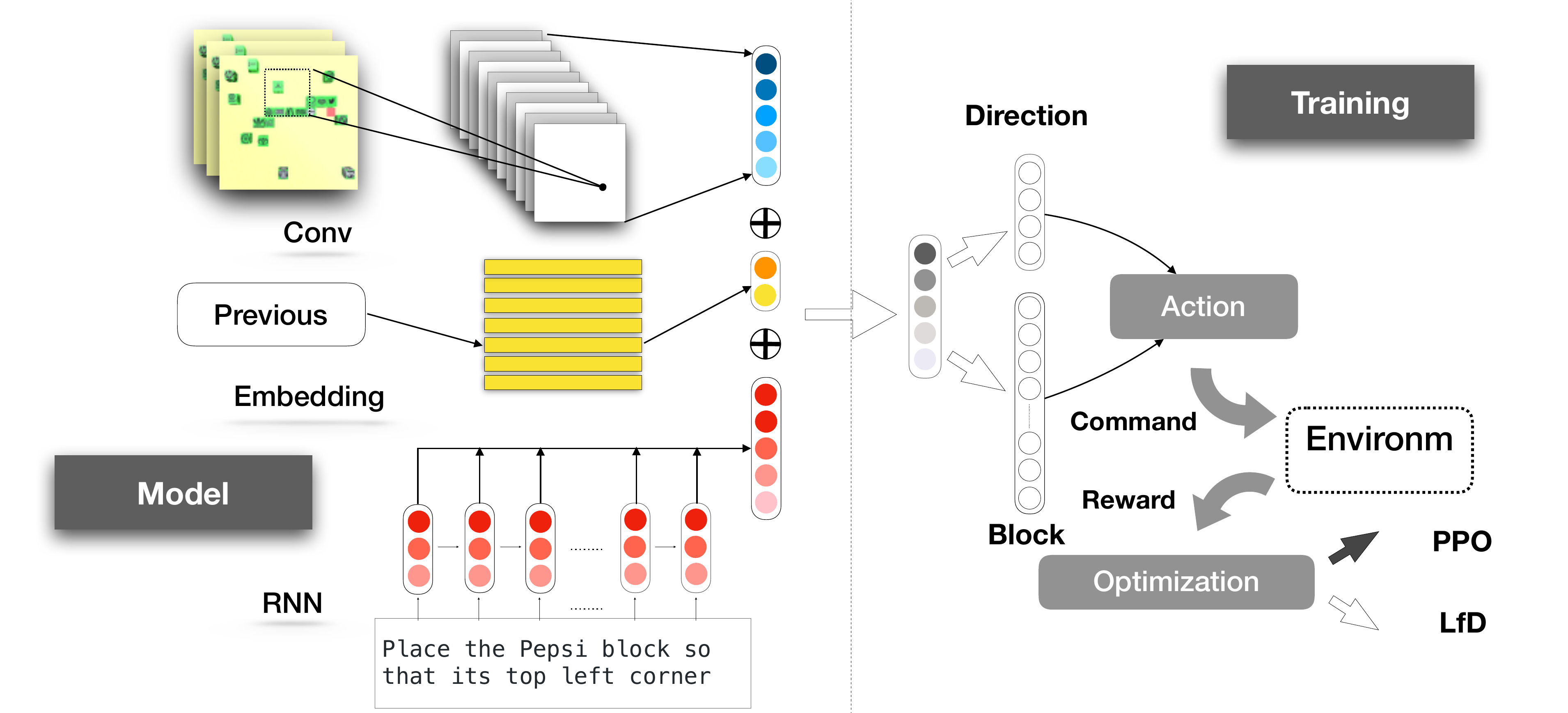}
\caption{Overview of our Scheduled Policy Optimization framework. The left part of the figure shows the structure of the policy network. The right part shows the Policy Optimization mechanism. The scheduler will keep track of the performance during training and maintain baseline value using moving averages. It alternately chooses between the RL update and LfD update.}
\label{method}
\end{figure*}

\section{Scheduled Policy Optimization for Natural Language Communication}
\label{sec:method}
\subsection{Task Formulation}

We consider an agent sequentially interacting with a block-world environment to accomplish a goal specified by a natural language instruction. 
For example, the agent may receive an instruction ``
\emph{move the block A to the right side of block B}''. The agent then moves certain blocks with a sequence of actions to accomplish the described task. 
Specifically, an instruction $ \mathbf{x} = \{ x_{1}, x_{2}, ..., x_{n} \} \in \mathcal{X}$ is a sequence of word tokens from a vocabulary $V$. At every time step, the agent perceives the environment state $o \in \mathcal{O}$ and outputs an action $a \in \mathcal{A}$.  The environment state could be a top-down view image of the map and the action could be ``\emph{move block-A north}''. Since the agent's action selection would depend on both  the given instruction and the environment state, we denote the joint of the instruction and the environment state as the state of the agent, $s=(\mathbf{x},o) \in \mathcal{S}$.\footnote{$\mathcal{S}=\mathcal{X} \times \mathcal{O}$} 
The agent's behavior is determined by a policy function $\pi: \mathcal{S} \times \mathcal{A} \rightarrow [0,1]$, which maps the  agent state into a distribution over actions. 

At time step $t$ the agent receives an immediate scalar reward $r_{t}$. The scalar is affected by the dynamics of the environment and the agent's actions. The goal of the agent is to find an optimal policy  maximizing the expected sum of discounted rewards,
\[
\max_{\pi} \mathcal{L^{\pi}} = \mathbb{E} \Big[ \sum_{t=1}^{\infty} \gamma^{t-1} r_{t} | \pi \Big]
\]
where $\gamma \in [0, 1)$ is a discount factor determining the tradeoff between short-term and long-term rewards. Deriving the optimal policy is practical via either learning from demonstration or reinforcement learning methods. 

The core of the agent is the policy. Since the agent states consist of instructions and environment states (images), a successful policy architecture thus should be able to handle both language understanding and grounding problems.

\subsection{Policy Architecture}

We use the same policy neural network architecture as \cite{misra2017mapping} for our agent. As depicted in Figure~\ref{method}, the policy architecture takes three inputs. The \textbf{environment state encoder} converts the images, $ o \in \mathbb{R}^{120\times 120 \times 3}$,  to a vector via convolutional neural networks, $s_{o}=\textrm{ConvNet}(o)$. The \textbf{instruction encoder} utilizes LSTM~\cite{hochreiter1997long} to encode the instruction. The word tokens $\{w_1,w_2,...,w_n\}$ are represented as one-hot vectors and then passed to a word embedding matrix $W^{I} \in \mathbb{R}^{D_{1}\times |V|}$, followed by the LSTM, i.e. $ h_i = \textrm{LSTM}(W^{I} w_i, h_{i-1}) $， where $D_{1}$ is the word embedding size and $|V|$ is the vocabulary size. The instruction sentence is then represented as the average of of the LSTM outputs, $s_{\textbf{x}}=\frac{1}{n}\sum_{i=1}^{n}h_{i}$. To avoid repeated failed actions, the last action $a$ is incorporated using an action encoding matrix $W^{A} \in \mathbb{R}^{D_{2}\times |A|}$, $s_{a}=W^{A}a$, where $D_{2}$ is the action embedding size and $|A|$ is the number of actions. The agent state $s$ is the concatenation of the visual, text and action vectors,  $s = s_{o} \oplus s_{\textbf{x}} \oplus s_{a}$. The agent state vector is passed through linear layers for predicting the two components of the action, where the first component is the \textbf{block ID} to move, and the second is the \textbf{movement direction}. Both are one-hot predictions.
\subsection{Scheduled Policy Optimization}

Direct reinforcement learning in a complex environment can be challenging especially when the state-action space is large. In our case, the agent only obtains the maximal reward when the instruction is accomplished and the probability of accomplishing the instruction via a random policy is exponentially decayed by a factor of 81, the number of the agent actions (4 directions $\times$ 20 blocks; and one special STOP action). Since the agent barely finds the optimal path during exploration, the training can be slow and ineffective.

To mitigate this problem, expert demonstrations are widely used to warm-start the initial policy.   \cite{misra2017mapping} 
collected a set of off-line demonstration to derive shaping reward to mitigate the delayed rewards. An orthogonal approach to leverage labeled expert actions is learning from demonstration, also referred as imitation learning or apprenticeship learning.
However, since the demonstrations are collected off-line in our problem, no supervision would be available when the agent's behavior is divergent from the demonstration. \cite{ross2011reduction} address the state distribution mismatch issue in LfD but their proposed method requires that demonstration must be collected on-line. 
We also observe that the learned policy from direct LfD has low entropy and thus it barely explores the environment, which makes the agent stuck at local minima. Brittle and tricky relaxation of the learned policy may introduce additional entropy, but it requires a significant amount of human tuning.  

In contrast to LfD, RL method is able to use its current policy to explore the environment and leverage new experience to bootstrap the policy. A combination of the LfD and RL could utilize the merits of both worlds. Our technical contribution of this paper is a new scheduled policy optimization algorithm which adaptively alternates between learning from demonstration and reinforcement learning.

\begin{algorithm}[t]
\small
\caption{Scheduled Policy Optimization Algorithm \label{algorithm}}
Randomly initialize policy network $\pi_\theta$;\\
Initialize learning history $\mathcal{H} \leftarrow \mathbf{\emptyset}$;\\
Scheduling flag $s_{\textrm{flag}} \leftarrow $ False;\\
\For{\textrm{epoch} $\leftarrow$ 1 \KwTo N}{
	\For{\textrm{sample} $\in$ \textrm{trainSet}}{
      Initialize episode length $steps \leftarrow 0$\\
        \If{$s_{\textrm{flag}}$}{
        	Retrieve expert trajectory;\\
            Append expert trajectory execution error $e$ to $\mathcal{H}$;\\
            Update $\theta$ using:
            $g \propto \nabla_\theta \mathcal{J}^{BC}(\theta)$;\\
            $s_{\textrm{flag}}\leftarrow $ False
        }
		\Else{
        	Sample action path $\tau \sim \pi_\theta$ until termination;\\
            $b = \textrm{average}(\mathcal{H}) + \lambda \sigma_c $\\
            Append execution error $e$ to $\mathcal{H}$;\\
            Update $\theta$ using:
            $g \propto \nabla_\theta \mathcal{J}^{PPO}(\theta)$;\\
            \uIf{$e > b$}{
            	$s_{\textrm{flag}} \leftarrow$ True
            }
        }
}
}
\end{algorithm}

\subsubsection{Schedule Candidates}
We investigate various schedule schema and reinforcement learning approaches in our experiments. Potential candidates of the scheduling mechanism are discussed as follows:

\paragraph{Deterministic Scheduling} The 
simplest is to schedule the LfD every $\mathcal{N}$ updates. This brings effective learning at the early stage because it mitigates sparse rewards. However, as the policy improves, the deterministic scheduler may fail to encourage sufficient exploration, which makes the training less efficient. 

\paragraph{$\mathbf{\epsilon}$-Sampling} As an simple improvement of the deterministic scheduling, the $\mathbf{\epsilon}$-Sampling is able to adapt the probability of LfD as training progresses by reducing $\epsilon$. The drawback is that adaptively setting $\epsilon$ requires lots of human hyper-parameter tuning.
\paragraph{History Baseline} The learning status of the agent could be measured via a windowed moving average of its execution performance $\mathcal{H}$. In our case we use the minimum number of steps from the final state of a trial to accomplish the instruction as the performance measurement. The larger the step number is, the worse the trial would be.  The LfD update is called to guide the learning progress whenever the last trial is worse than the baseline estimate:\[
b = \textrm{average}(\mathcal{H}) + \lambda \sigma_c 
\]
where $\sigma_c$ is the standard error of the mean estimate and the 
 coefficient $\lambda > 0$ is a hyper-parameter controlling the convergence speed. 
This schedule schema is able to adaptively utilize the imitation learning and allows more RL exploration. The schedule schema will call LfD less as the learning progresses because it becomes less likely for the agent to be worse than the baseline.

Our best model is based on the baseline scheduler coupled with PPO algorithm, which is less sensitive to hyperparameters. Since the baseline module uses an adaptive baseline estimator which measures the policy's real time performance, it tends to give more consistent improvements. Besides, PPO can provides a more stable baseline value compared to unconstrained policy gradient.

In our experiments, the empirical performance is optimal when the history baseline module is used. The pseudo code of our Scheduled Policy Optimization is shown in Algorithm~\ref{algorithm}. The policy learning algorithms we use are discussed below.

\subsubsection{Behavior Cloning}
As for LfD, we utilize Behavioral cloning~\cite{pomerleau1991efficient}, which is a widely used imitation learning approach. Its learning objective is to maximize the log likelihood of the demonstration actions:
\begin{equation*}
\begin{split}
\mathcal{J}^{BC}(\theta) = \frac{1}{N} \sum_{i=1}^{N} \log \big(\pi_\theta(a^*_i|s_i) \big)
\end{split}
\end{equation*}
where $\{(s_{i}, a^{*}_{i})\}_{i=1}^{N}$ is a set of demonstration state-action pairs and $\theta$ are the learnable parameters of the policy neural network. 

\subsubsection{Proximal Policy Optimization}
To obtain a stable baseline of execution performance, we use a recently proposed conservative policy gradient method, Proximal Policy Optimization (PPO)~\cite{schulman2017proximal}, as our RL algorithm. PPO defines a surrogate objective which is the lower bound of the true reward objective: 
\begin{align*}
\mathcal{J}^{PPO}(\theta) &= \mathbb{E}\Big[\min \big(\rho_t(\theta)A_{t},  [\rho_t(\theta)]^{1+\epsilon}_{1-\epsilon} A_{t})\Big] \\
\rho_t(\theta) &= \frac{\pi_\theta(a_t|s_t)}{\pi_{\theta_{\textrm{old}}}(a_t|s_t)} 
\end{align*}
where $[.]^{a}_{b}$ is a clip function in the interval $[a, b]$ and $A_{t} = R_t - V(s_t)$ is the advantage function, calculated as the difference between reward and the state value estimate of time step $t$. The state values estimator is learned by minimizing the mean square error between $r_t$ and $V(s_t)$:
\begin{align*}
\mathcal{J}_{value}(\theta) &= \mathbb{E}[(r_t - V(s_t))^2]
\end{align*}
When compared to directly optimizing the reward objective $\mathbb{E}[A_t]$, optimizing this lower bound $\mathcal{J}^{PPO}$ can better guarantee monotonic policy improvements.

\section{Experiments}
\label{sec:exp}

\subsection{Dataset}
We evaluate our scheduled policy optimization method on the Blocks environment originally created by Bisk \emph{et al.}~\shortcite{bisk2016natural}. 
There are 20 unique blocks in the environment and the goal of the agent is to accomplish natural language described tasks by moving blocks in the 2D map. The dataset consists of 11,871 training samples and 1,179/3,177 samples for validation/testing. To speed up training, previous work applied reward shaping techniques in designing immediate rewards based on the environment's internal states. To make the results comparable we use the same reward functions as Misra \emph{et al.}~\shortcite{misra2017mapping}. The performance of the learned policies is measured by the execution error, which is the minimum number of steps to accomplish the task from the last state in a trial. The lower the execution errors are, the better the learned policy would be. Note that Misra \emph{et al.}~\shortcite{misra2017mapping} also report the minimum distance metric. As the released simulator does not provide this number, we are unable to compare this metric.

\subsection{Training Details}
Our model is implemented using PyTorch~\cite{paszke2017automatic}. We use Adam optimizer~\cite{kingma2014adam} to update the model parameters. The initial learning rate is $0.0001$ and is divided by 2 for every 4 epochs. The windowed history consists of the execution errors of the last 100 trials. The clipping interval of PPO is set to $[0.95,1.05]$ and the number of PPO epochs for each update step is set to be 4. We restrict the number of training epochs to be less than 20. Early-stopping is applied using the Dev set.\footnote{Code and trained models can be found at \url{https://github.com/xwhan/walk_the_blocks}.} In addition to the PPO algorithm, we also include the results of using other reinforcement learning algorithms, REINFORCE~\cite{williams1992simple} and advantage actor-critic (A2C)~\cite{peters2008natural}, to demonstrate the general improvements from the scheduled policy optimization schema. In order to achieve more stable training, entropy regularization with the same coefficient (0.1) is also added to all these models.

\subsection{Baselines}
We include results from ~\cite{misra2017mapping} as baselines:

\noindent \textbf{HUMAN} is human demonstration. It is also the lower bound of the performance.

\noindent \textbf{INITIAL} is the agent taking no actions and the trial terminates at the initial state. It can also be viewed as the average distance between the initial state and the goal state. 

\noindent \textbf{RANDOM} is the agent taking random actions. Note that it is generally worse than the INITIAL baseline because the random actions even increase the average distance. 

\noindent \textbf{Ensem-LfD} is trained via learning from demonstration only. Trained models are ensembled for better performance. 

\noindent \textbf{Ensem-DQN} is trained using reward shaping techniques via DQN. No demonstration is used to initialize the network.

\noindent \textbf{Ensem-REIN} is initialized with supervised learning from demonstrations and then retrained by REINFORCE algorithm using cumulative rewards.

\noindent \textbf{Ensem-BEST} is initialized with supervised learning from demonstrations and then retrained by REINFORCE algorithm using shaped intermediate rewards.  

\subsection{Main Results}
\begin{table}
\centering
\small
\begin{tabular}{lc|c|c|c}
\toprule
& \multicolumn{2}{c}{Dev Error} &\multicolumn{2}{c}{Test Error}\\
\cmidrule{2-5}
Methods & Mean & Med. & Mean & Med.\\ \midrule
\textsc{Human} & 0.35 & 0.30 & 0.37 & 0.31\\ \midrule
\textsc{Initial} & 5.95 & 5.71 & 6.23 & 6.12\\
\textsc{Random} & 15.3 & 15.70 & 15.11 & 15.35\\ \midrule
\textbf{Misra el al.}  \\
Ensem-LfD & 4.64 & 4.27 & 4.95 & 4.53\\
Ensem-DQN & 5.85 & 5.59 & 6.15 & 5.97\\
Ensem-REIN & 5.28 & 5.23 & 5.69 & 5.57\\
Ensem-BEST & 3.59 & 3.03 & 3.78 & 3.14\\
\midrule 
\textbf{Our Models}\\
S-REIN & 2.94 & 2.23 & 2.95 & 2.21\\
S-A2C & 2.79 & 2.21 & 2.75 & 2.18\\
S-PPO & \textbf{1.69} & \textbf{0.99} & \textbf{1.71} & \textbf{1.04}\\
\bottomrule
\end{tabular}
\caption{Performance (mean and median of execution errors) of our scheduled policy optimization and baselines. The numbers of the baselines are from \protect\cite{misra2017mapping}.}
\label{result2}
\end{table}
Table~\ref{result2} summarizes the performance of our agents and the baselines on the dev and test sets. The performance is measured as the minimal number of steps from the final state of a trial to accomplish the instruction (final distance from the target). We denote our agent as Scheduled $X$ (S-$X$), where $X$ could be REINFORCE, advantage actor-critic (A2C) or PPO.  Note that instead of using ensembles to achieve best results all of our agent results are generated using single models.

Our scheduled policy gradient variants (S-REIN/S-A2C/S-PPO) have significant lower errors than the best baseline (Ensem-BEST). S-PPO (scheduled PPO) is able to move the blocks to the positions that are only $1\sim2$ blocks away from the goal locations while Ensem-BEST can only move to $3\sim4$ away locations. Consider the initial distance is around 6 blocks, our scheduled policy optimization methods make substantial progress. 

We also notice that the scheduled systems with unconstrained policy gradients generate very similar performance while the scheduled PPO is able to give much better result, potentially due to its stable policy updates and accurate baseline values.

The performance of Ensem-LfD also suggests that learning from demonstration approach fails to generalize in this task. Compared to the INITIAL baseline, the improvement of Ensem-LfD is limited to only one block.

\subsection{Analysis of the Exploration Strategies}

The exploration strategy plays an essential role in the agent's policy learning. 
Insufficient exploration can lead to a local-optimal policy which may not generalize well during test, while too much exploration can be significantly inefficient. 
Ideally, we want the agent to do effective exploration during early stage of training; as training goes on, it should be able to converge to a near-greedy policy. We show that our schedule policy optimization has indeed achieved this kind of exploration strategy.

We compare our schedule policy optimization method (S-PPO) to two baselines, a pure PPO method (PPO) and a naive combination of LfD and PPO (LfD-PPO). LfD-PPO initializes the policy using LfD and then use PPO to fine-tune the model.  

\begin{figure}[t]
\centering
\includegraphics[width=0.9\linewidth]{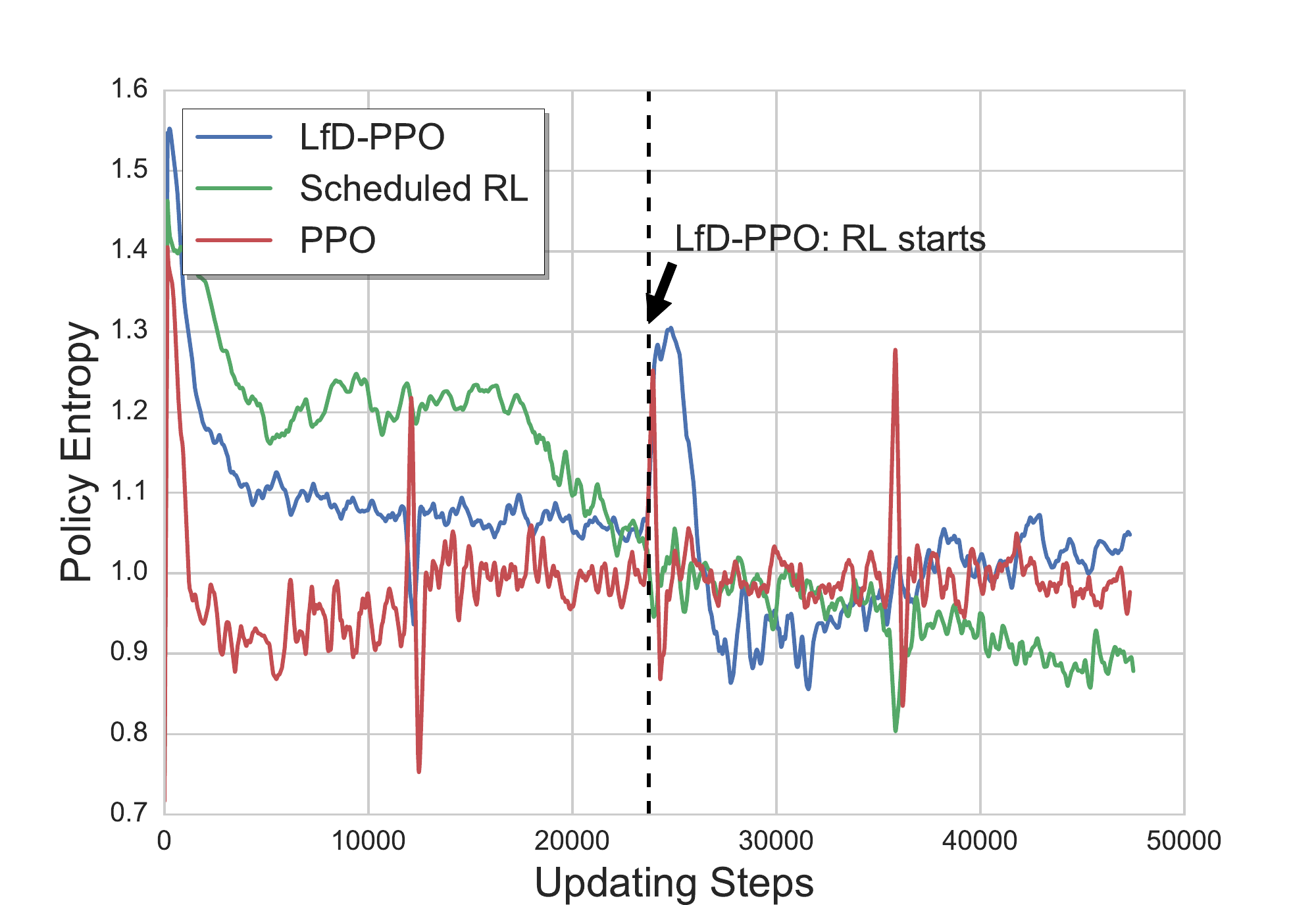}
\caption{Curves of policy entropy in training for the three agents. The spikes in the curve are caused by random shuffle after each epoch. \textbf{LfD-PPO}: PPO with supervised learning for initialization; \textbf{PPO}: A pure RL algorithm with PPO updates only; \textbf{Scheduled RL}: our proposed method.}
\label{entropy}
\end{figure}

\begin{figure}[t]
\centering
\includegraphics[width=0.9\linewidth]{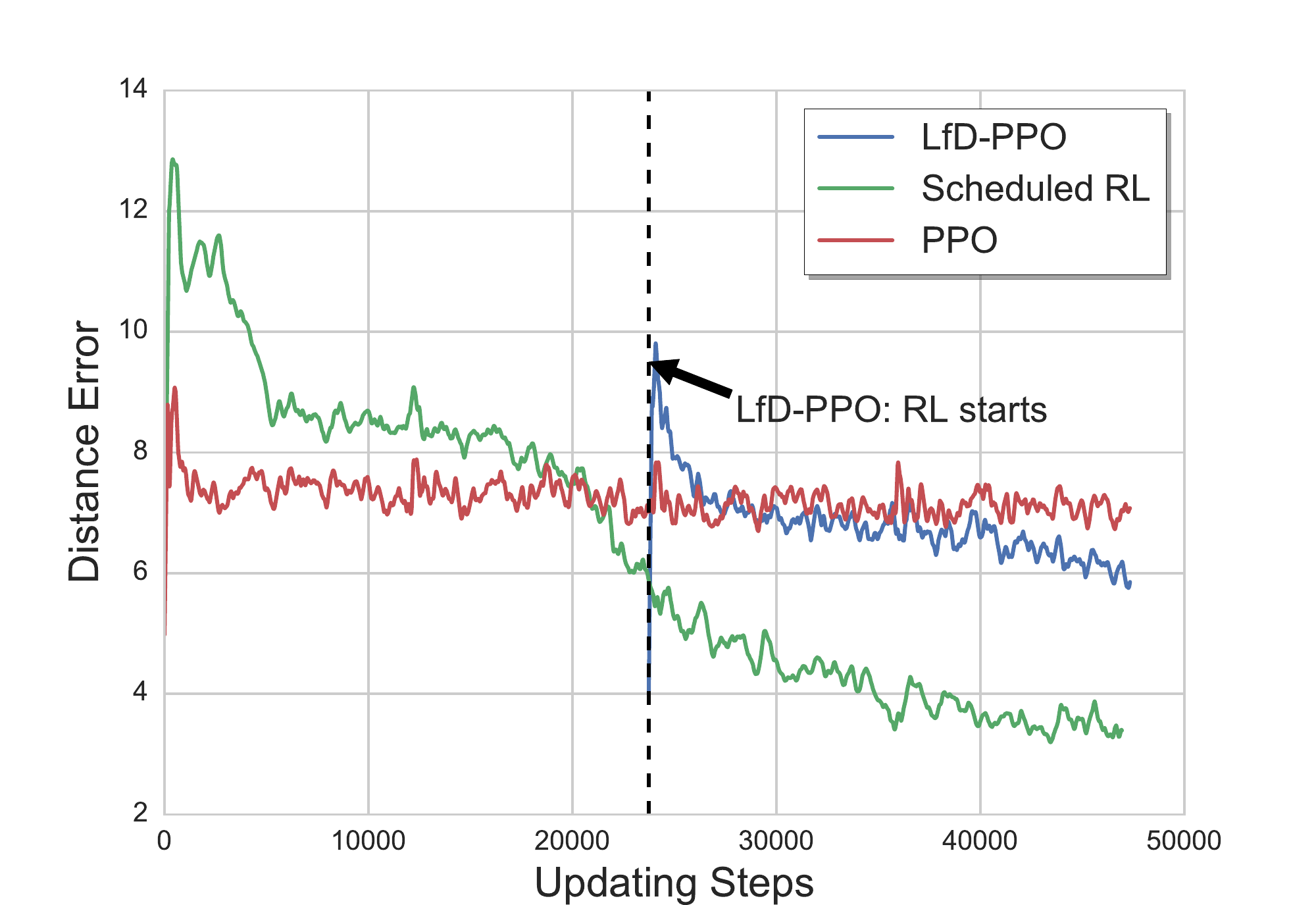}
\caption{Curves of distance errors in training for the three agents. Only the distance errors in the PPO re-training of the \textbf{LfD-PPO} baseline are shown.}
\label{error}
\end{figure}

We keep track of the policy entropy (Figure~\ref{entropy}) and the policy performance (Figure~\ref{error}) in learning. We examine the policy entropy because it is a good indicator of the policy exploration. Since entropy indicates the randomness of a distribution, a high-entropy policy tends to explore the surrounding area of the greedy paths while the policy with a low entropy usually sticks to the greedy path.
The policy performance is measured by error curves Figure~\ref{error}, which show the execution errors of different agents during training. The error curves reflect the efficacy of learning. 
We notice significant difference in the exploration strategies of the three agents.

The PPO agent's policy entropy decreases dramatically at the beginning of training but its execution performance fails to improve.
Our hypothesis is that the PPO agent is vulnerable to suboptimal policies. Once the agent obtains much better rewards than previous encountered rewards, the update gradient will push the distribution density to this particular action. Since the actions are sampled from the distribution, it becomes more likely that the agent will sample the same action again and get an update gradient in the same direction. This might lead to a low-entropy but sub-optimal policy.

As for the LfD-PPO agent, the supervised learning may also result in a relatively low-entropy policy. When the PPO training starts, the agent maintains a high entropy for a short time and then quickly goes back to a low-entropy policy. This learning pattern indicates a defective policy that fails to do sufficient exploration in the environment, leading to slow training, as indicated by Figure~\ref{error}.

Compared to both the PPO agent and LfD-PPO agent, our RL agent with the scheduling mechanism has a higher entropy at the early stage of training. As training goes on, the policy is able to converge to a low-entropy distribution. Also, we can see from Figure~\ref{error} that the learning process of our agent is also much more effective than the other baselines.

\subsection{Comparison with LfD}
Compared to LfD, our method has two major advantages.

\begin{figure}
\centering
\includegraphics[width=0.9\linewidth]{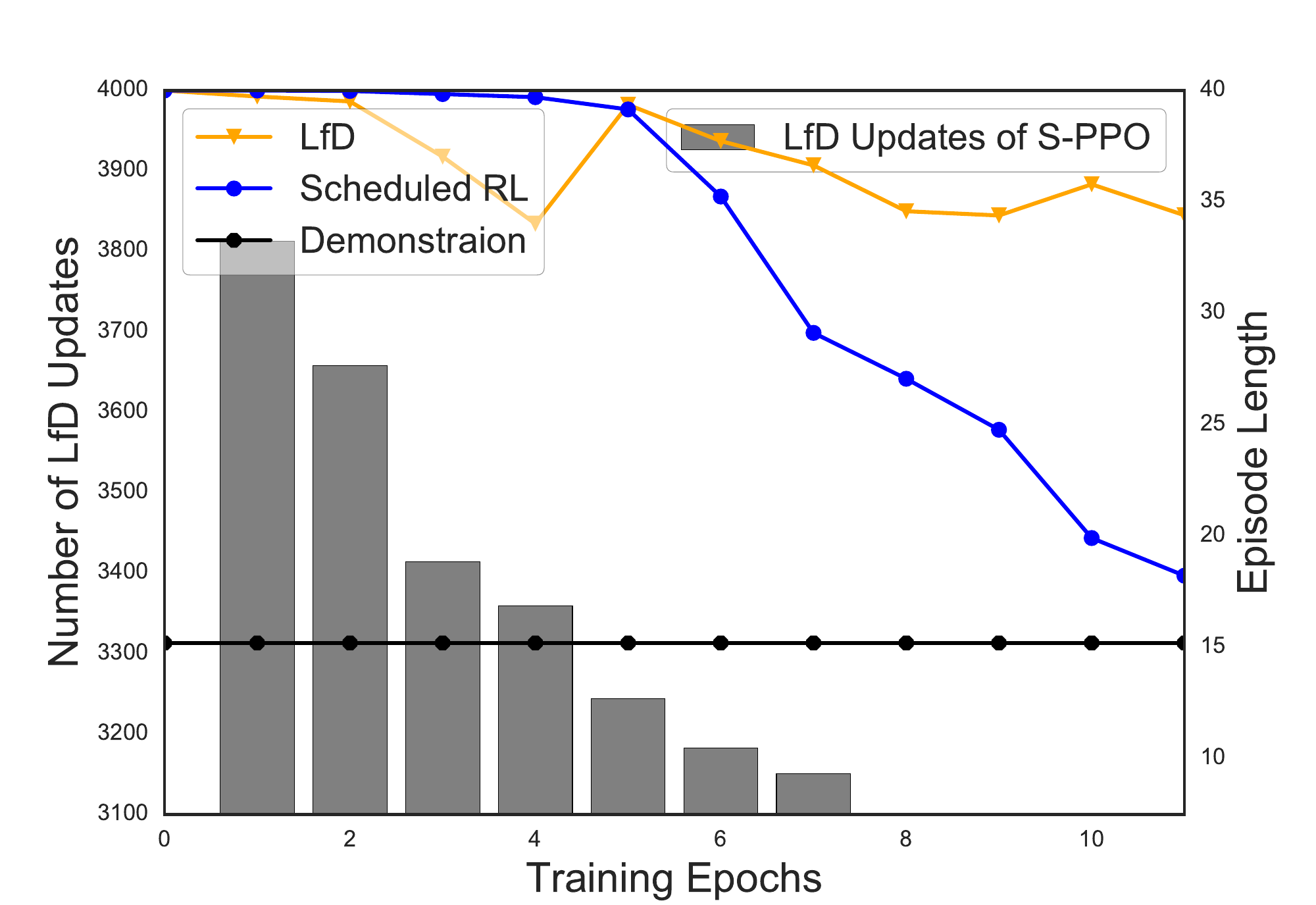}
\caption{Comparison between LfD and our Scheduled RL in terms of number of LfD updates and episode length. The bars show the number of LfD updates in each training epoch for the Scheduled RL. The lines are the average episode lengths on the dev set.}
\label{length}
\end{figure}

First, fewer demonstrations data are required to learn a generalizable policy. As shown by the bar plot in Figure~\ref{length}, our scheduled RL agent uses fewer and fewer demonstrations during training. This also indicates consistent policy improvement. Note that for LfD, all the demonstrations are used in every epoch.

Next, we look into the episode lengths during inference to see if the agent is able to finish the task efficiently. As indicated by Figure~\ref{length}, which shows the average episode lengths (steps of actions) on the dev set after every training epoch, the LfD agent often exhausts the maximum action steps. However, we notice that our scheduled RL can alleviate this problem as we conduct more training epochs. The black line shows the average lengths of human demonstrations, which serve as the baseline. At the early training stage, the average length is close to 40, which is set as the maximum action steps. As training goes on, the episode lengths of our scheduled agent is getting closer to the average demonstration length. In contrast, The supervised model still fails to output \textsc{STOP} after a long training time. One possible explanation for this phenomenon is related to the class imbalance problem of machine learning. In the supervised settings, every state-action pair is used as one training sample. It is obvious that there are much more non-stop actions than stop actions, which makes the labels largely imbalanced. This makes it rather difficult for the LfD agent to recognize target states. Whereas for RL, as it collects training data by sampling from the policy, it can be vulnerable to imbalanced classes.

\section{Related Work}
\label{sec:related}

The task of learning to understand free-form instructions has attracted lots of attention since early stage of AI~\cite{di1992understanding,winograd1972understanding}. MacMahon \emph{et al.}~\shortcite{macmahon2006walk} build a system based on linguistic and execution modules. Their model requires both spatial and linguistic prior knowledge and cannot be trained end-to-end. Tellex \emph{et al.}\shortcite{tellex2011understanding} develop an approach based on probabilistic graphical models. Their approach requires a semantic map of the environment, which may not be available for complex environments. Some recent studies~\cite{kim2013adapting,mei2016listen} assume no prior linguistic knowledge and formulate the task as an encoder-decoder problem, where free-form texts are directly mapped into executable actions. These models take simple discrete state inputs while our model is able to take raw RGB images as inputs.

Although the problem of instruction understanding has been extensively studied, only a few methods take into account the state change of the environment during execution. Branavan \emph{et al.}~\shortcite{branavan2009reinforcement} are the first to apply RL to learn a mapping between documents and the sequence of actions, which considers the state transition dynamics. However, their method is based on a simple log-linear model, which is also hard to generalize to multi-modal state inputs. On the other hand, with the success of deep reinforcement learning (DRL)~\cite{mnih2015human,silver2017mastering},  Misra \emph{et al.}~\shortcite{misra2017mapping} propose to model the action decoding as a Markov Decision Process using deep neural networks. Their model makes use of both human demonstration actions and shaped rewards for training. The authors test various RL algorithms, however, the performance is still far from human performance. An earlier work~\cite{walsh2011blending} has explored the scheduling of imitation learning and RL but the authors make much stronger assumption about the coverage of demonstration actions. While they require the demonstration to cover the whole action space, our method only needs a fixed set of demonstrations. 

\section{Conclusion}
\label{sec:con}
We study the problem of directly mapping human language instructions and raw image observations into effective action sequence. On the Blocks environment, the proposed RL framework outperforms the existing methods by $55\%$ in terms of exexution error. Compared to existing methods which use human demonstration to pre-train the network, our scheduling mechanism takes both generalization and data efficiency into account. By utilizing an adaptive scheduling mechanism which alternates between LfD (imitation learning) and conservative policy updates, the RL agent is able to maintain a high-entropy training policy for sufficient exploration without sacrificing the learning efficiency. Besides, since there is no extensive pre-training in our framework, much fewer demonstration paths are required to train our model.

\section*{Acknowledgments}
We are grateful for the support of an IBM Faculty Award.

\bibliographystyle{named}
\bibliography{ijcai18}

\appendix
\section{Additional Experiments on A New Dataset}
\begin{figure}[h]
\centering
\includegraphics[width=1.0\linewidth]{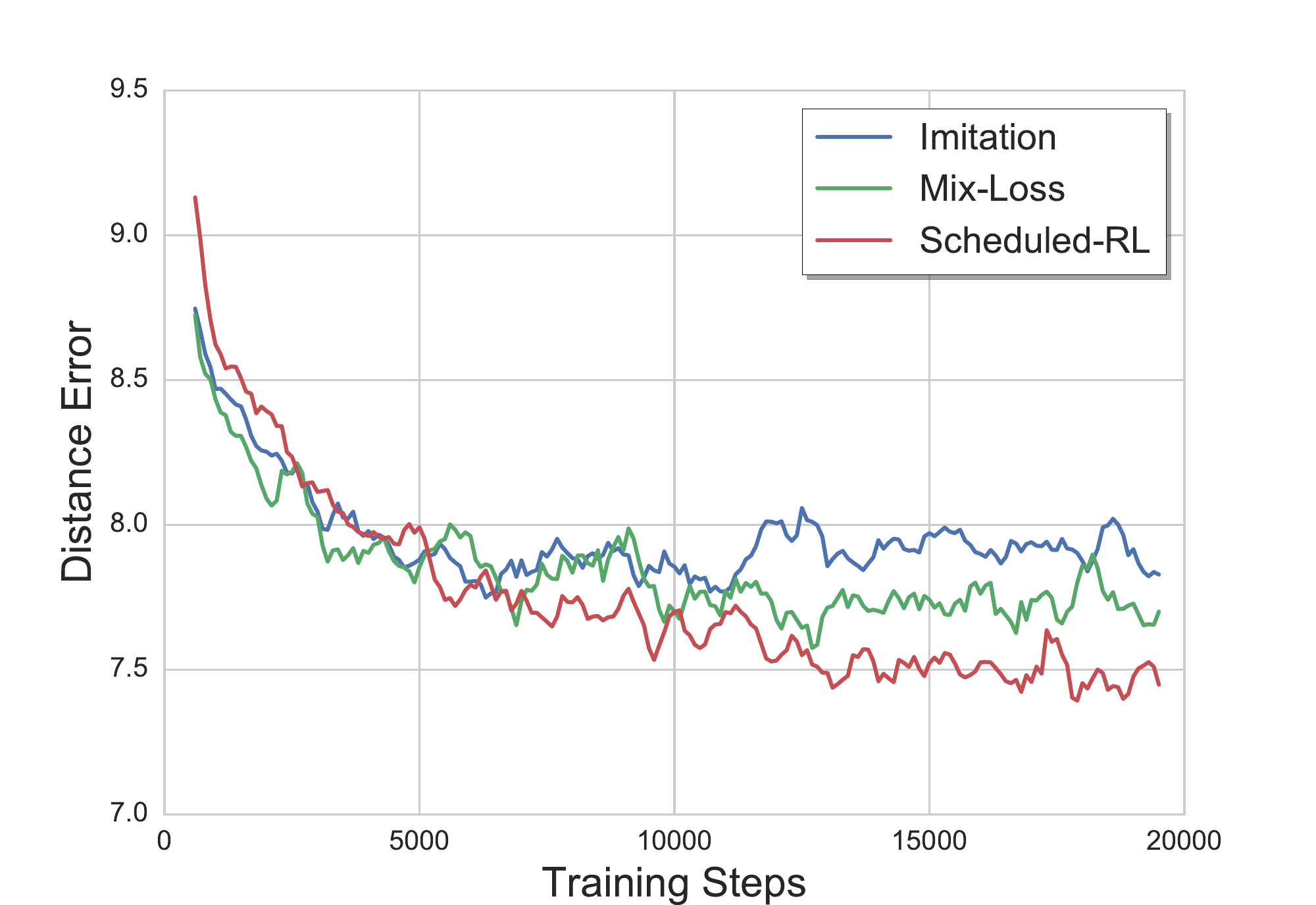}
\caption{Distance error evaluated on the unseen development scenes of Room-to-Room environment.}
\label{addtion}
\end{figure}

More recently, a new dataset~\cite{anderson2017vision} with realistic indoor scenes has been released. This dataset (Room-to-Room) includes 21,567 crowd-sourced natural language instructions and 10,800 panoramic RGB-D images. To the best of our knowledge, this is the first instruction-following dataset that is made of real images. To show that the scheduled mechanism is able to provide general improvements, we compare our scheduled RL with vanilla RL and a mix-loss~\cite{ranzato2015sequence} method on this dataset. We use a similar network architecture as in~\cite{anderson2017vision}. Instead of training the agent using only a cross-entropy loss to imitate demonstration actions, we introduce a distance-based reward. We conduct some initial experiments using the released development environment, which includes only unseen scenes. Figure~\ref{addtion} shows the curves of distance error (distance between the agent's final position and the target position) calculated on the unseen scenes from the development set. We can see that the proposed scheduled RL algorithm is superior to both cross-entropy training and mixed-loss training. However, we also notice that on the seen scenes, our RL algorithm does not provide much improvement. We leave further investigation on this dataset to future work.

\end{document}